\documentclass{llncs}
\usepackage{amsmath}
\usepackage{amsfonts}
\usepackage{amssymb}
\usepackage[figuresright]{rotating}
\usepackage{array}
\usepackage{setspace}
\usepackage{url}
\usepackage{multirow}
\usepackage{latexsym}
\usepackage{amsmath,amssymb,amsfonts}
\usepackage[utf8]{inputenc}
\usepackage[T1]{fontenc}
\usepackage[english]{babel}
\usepackage{verbatim}
\usepackage{booktabs}
\usepackage{xspace,xcolor}
\usepackage{graphicx}
\usepackage{soul}
\usepackage{makecell}
\usepackage{csquotes}
\usepackage{pgfplots}

\usepackage{pgf}
\usepackage{pdfpages}
\usepackage{float}
\usepackage{amsmath}

\usepackage{enumerate}
\usepackage{fixltx2e}
\usepackage{cite}
\usepackage{subcaption}
\urldef{\mails}\path|dr.rajivbajpai@gmail.com, {hazarika,rogerz}@comp.nus.edu.sg,|
\urldef{\mailss}\path|kunal.singh@iitkgp.ac.in, gorantlas@iisc.ac.in,|
\urldef{\mailsss}\path|cambria@ntu.edu.sg|

\newcommand*\samethanks[1][\value{footnote}]{\footnotemark[#1]}    
\begin{document}
	\pagestyle{empty}
    
	\title{Aspect-Sentiment Embeddings for Company Profiling and Employee Opinion Mining}
	\author{Anonymous}
    \institute{  }
	\author{Rajiv Bajpai\thanks{Equal contribution.}\inst{1} \and Devamanyu Hazarika\samethanks\inst{2} \and Kunal Singh\inst{4} \and Sruthi Gorantla\inst{5} 
   \and Erik Cambria\inst{3} \and Roger Zimmermann\inst{2}}
	\institute{Accenture, India\\
		\and National University of Singapore, Singapore\\
		\and Nanyang Technological University, Singapore\\
		\and Indian Institute of Technology Kharagpur, India\\
		\and Indian Institute of Science, Bangalore, India
	\begin{center}
        \small{
		\mails\\
        \mailss\\
		\mailsss\\}
		\end{center}
		}
	
	\maketitle

	\begin{abstract}
	With the multitude of companies and organizations abound today, ranking them and choosing one out of the many is a difficult and cumbersome task. Although there are many available metrics that rank companies, there is an inherent need for a generalized metric that takes into account the different aspects that constitute employee opinions of the companies. 
In this work, we aim to overcome the aforementioned problem by generating aspect-sentiment based embedding for the companies by looking into reliable employee reviews of them. We created a comprehensive dataset of company reviews from the famous website \url{Glassdoor.com} and employed a novel ensemble approach to perform aspect-level sentiment analysis.
	Although a relevant amount of work has been done on reviews centered on subjects like movies, music, etc., this work is the first of its kind. We also provide several insights from the collated embeddings, thus helping users gain a better understanding of their options as well as select companies using customized preferences.
	\end{abstract}

	\section{Introduction}
	Emotions, sentiment, and judgments on the scale of good---bad, desirable---undesirable, approval---disapproval are essential for human-to-human communication.
	Understanding human emotions, deciphering humans' emotional reasoning and how humans express them in their language is key to enhancing human-machine interaction. In this era of social media, the WWW provides new tools that create and share ideas and opinions with everyone efficiently. 
Capturing public opinion on social media about events, political movements or any other topics bears a potential for interest amongst the scientific community. That is mainly because of two reasons - firstly, these opinions can help individuals in their decision making process. Secondly, organizations will utilize such data in order to glean public opinion regarding  services and products so as to fine-tune/develop their business strategies. 
	
	Sentiment analysis is the study of opinions and sentiments from content that spans from the unimodal to the multimodal \cite{shah2016multimodal,shah2014advisor,shah2016leveraging}. Recent approaches to sentiment analysis have focused on the use of linguistic patterns and deep neural networks. The ability to identify aspects within texts is also equally important. An aspect is defined as the product feature; for instance, in the sentence ``the battery lasts long'', \emph{battery} is the aspect for which positive sentiment is expressed. The main challenge of sentiment analysis is identifying aspects and their corresponding sentiment. In our case, we do so by merging linguistic patterns and an ELM classifier.
	
	Employees are organizational assets and their opinion plays a vital role in any organization's growth. Job Search Engines and review websites have evolved to become an ocean of employee reviews. Employee reviews play a vital role for a company's growth as it improves the relationship between management and the employees via improving staff welfare and morale. These reviews also help prospective employees in selecting a company that meets their criterion. Despite such reviews being important data sources for sentiment mining, they have failed to draw the attention of the scientific community. To the best of our knowledge, only the work of \cite{moniz2014sentiment} utilized company reviews from Glassdoor. However, even their work was limited to extracting only topics and sentiments from the reviews. In this work, we built a large dataset of employee reviews of companies in Singapore sourced from Glassdoor. Different types of analysis, e.g., aspect extraction, aspect based sentiment analysis were then carried out on this dataset by blending ELM with sentic patterns. To this end, we developed representational embeddings of the companies based on the sentiment score of various different aspects of the companies. In particular, each company is represented in a 30 dimensional space where each dimension corresponds to the average sentiment score of an aspect. Some of these aspects are `company culture', `salary', `location', etc.
	
	In this paper, we used Glassdoor as our source for the preparation of the dataset comprising of 40k reviews. The volume of reviews span a diverse range of aspects (positive and negative) that describe the company based on personal opinions of the reviewers. There are two main contributions of this paper:
	
	\begin{itemize}
		\item Creation of a large dataset derived from Glassdoor for aspect level sentiment analysis. This dataset contains the reviews of employees working or who previously worked at the corresponding companies.
		\item Introducing aspect-sentiment embeddings of the companies in order to find similarities between companies in the similar or differing sectors. Aspect-sentiment embeddings project each company onto an n-dimensional space where each dimension corresponds with the overall aspect-sentiment strength of the employees. Aspects are different features of a company, e.g., \emph{salary, location, work-life balance}, etc. This is particularly useful for job seekers who are looking to find companies that suit their preferences.
	\end{itemize}
	
	The rest of the paper is organized as follows: Section \ref{sec:rel} discusses sentiment analysis literature; collection and preparation of the dataset are featured in Section \ref{sec:dataset}; we discuss the algorithm details in Section \ref{sec:algo}; experimental results of this study are presented in Section \ref{sec:exp}; finally, Section \ref{sec:con} concludes the paper.
	
	\subsection*{Related works }
	\label{sec:rel}
	Identifying emotions associated with employers is just one of the many possible applications of sentiment analysis. We could also analyze industries or professions as a whole, or consider the relationship between the emotional content of reviews with the corresponding salaries of employees. One would expect higher salaries to correlate with more positive emotions, but we might also see an inverse correlation in some cases, perhaps indicating the use of `golden handcuffs'.
	
	Sentiment analysis systems can be broadly categorized into knowledge-based \cite{camkno} or statistics-based systems \cite{camsta}. Initially, knowledge bases were more commonly used for the identification of emotions and polarity in text. However, at present, sentiment analysis researchers more commonly use statistics-based approaches, with a specific focus on supervised statistical methods. For example, Pang et al.~\cite{pang2002thumbs} compared the performance of different machine learning algorithms on a movie review dataset: using a large number of textual features, they obtained 82.90\% accuracy.
	
	Other unsupervised or knowledge-based approaches to sentiment analysis include Turney et al.~\cite{turney2002thumbs}, which used seed words to calculate the polarity and semantic orientation of phrases, as well as Melville et al.~\cite{hu2013unsupervised} which proposed a mathematical model to extract emotional clues from blogs and then used the information for sentiment detection.
	
	Sentiment analysis research can also be categorized as single-domain \cite{pang2002thumbs} versus cross-domain\cite{blitzer2007biographies}. The work presented in \cite{pan2010cross} discusses the use of spectral feature alignment to: 1) group domain-specific words from different domains into clusters, and 2) reduce the gap between domain-specific words of two domains using domain independent words. Bollegala et al.~\cite{bollegala2013cross} developed a sentiment-sensitive distributional thesaurus by using labeled training data from source domain and unlabeled training data from both source and target domains. Some recent approaches~\cite{denecke2008using,ohana2009sentiment} used SentiWordNet~\cite{baccianella2010sentiwordnet}, a very large sentiment lexicon developed by automatically assigning polarity value to WordNet~\cite{miller1995wordnet} synsets. In SentiWordNet, each synset has three sentiment scores along three sentiment dimensions: positivity, negativity, and objectivity.
	
As discussed in the introduction, there are hardly any works on mining opinions from company reviews written by employees. Moniz et al.~\cite{moniz2014sentiment} proposed an aspect-sentiment model based on the Latent Dirichlet Allocation (LDA). According to their study, the results of the articulate aspect-polarity model showed that it might be advantageous for investors to combine an appraisal of employee satisfaction with other existing methods for forecasting firm earnings. The research explained and analyzed the sentiments of a stakeholder group which is possibly neglected: the firm's employees. The researchers initially used online employee reviews in order to capture employee satisfaction and utilized LDA to consider salient aspects in employees' reviews. From that, they manually derived a latent topic that appeared to be associated with the firm's outlook.
	Secondly, they created an entire document by grouping employee reviews for each firm, and using the General Inquirer dictionary to count positive and negative terms, they measured sentiment as the polarity of the composite document. Their model suggested that employee satisfaction could be formulated as a function of the firm's outlook and employee sentiment.
	
	\section{Dataset collection}
	\label{sec:dataset}
	
	In our research, we used the popular job recruiting site \textit{Glassdoor.com} as our source to prepare the dataset. The website provides tons of reliable reviews for many companies written mostly by employees, ex-employees or directly associated clients. The content of the reviews possess different aspects (positive and negative) that represent the company from the individual perspectives of the writers. The anonymity of writers enhances the authenticity of the review, thus, this site was our primary source for the dataset collection. The language used in the reviews was found to be highly formal with very minimal usage of slang, decreasing the effort required for data cleaning, normalization, and tokenization. These `clean' words had a high match rate with the dictionary we used \cite{esuli2006SentiWordNet} for creating the word embeddings, thus improving the performance of the ELM model used in our ensemble network.
	
	The review structure in \url{Glassdoor.com} consists of a general description followed by both \textit{pros} and \textit{cons} to be written and listed by the writer. This rigid structure aids us in building a balanced dataset comprising of both positive and negative reviews associated with the companies. However, one must be wary of comments like ``I really don't have anything to say/complain about here'' in the pros/cons section, which we believe to represent false positives or false negatives respectively. We decided to include these comments in our dataset to represent real-world instances where such false comments are prevalent. 
	
	The dataset has been collected by using the official API\footnote{https://www.glassdoor.com/developer/index.htm} of Glassdoor. 
	We created a diverse list of 60 well-known companies representing various domains such as technology, finance, energy, hospitality, etc. Finally, we collected a total of 20,000 reviews for all companies. As mentioned earlier, the reviews listed both the pros and cons about the company that is reviewed. Given this sophistication, it was easier for us to split each review into two sub-reviews containing positive and negative opinions respectively. This in turn enabled the automatic labeling of said reviews. Despite doing so, we were careful to manually check the labeling afterwards in order to filter out wrong labels. 
	
	Here is an excerpt from one of the positive reviews written for \textit{Accenture}-
	\textit{``They have great career opportunities, a never ending supply of interesting work, competitive compensation, wonderful benefits, great people, wonderful training programs, a tremendous number of brilliant professionals in their fields ready to help, and great core values''}. This review, like others, is full of important aspects such as \textit{opportunities, compensation, benefits, etc.}, which provides extensive opinions on different facets/characteristics of the company and thus allow our team to prepare a comprehensive review dataset.
	
    \begin{table}[h]
		\centering
        \begin{scriptsize}
		\begin{tabular}{|c| c| c| c |} 
			\hline
			Company &  Reviews & Company &  Reviews  \\ [0.5ex] 
			\hline\hline
			Accenture & 1000 		& HP & 150 \\
			Adobe & 998 			& HSBC Holdings & 1850 \\
			Aeropostale & 874 		& IBM & 150 \\
			Aflac & 368 			& Intel Corporation & 998 \\
			Autodesk & 752 		& Intuit & 972 \\
			Bank of China & 212 		& Marriot International & 980 \\
			Booz Allen Hamilton & 976    & Microsoft & 1000 \\
			Broadcom & 151 		& Mosanto & 396 \\
			Brocade & 990 		& Morningstar & 733 \\
			Camden Property & 203 	& National Instruments & 712 \\
			Capital One & 997 		& NetApp & 800 \\
			CarMax & 959 		& Nordstorm & 839 \\
			Chesapeake Energy & 725    & OCBC & 268 \\
			Cisco & 980 			& Paychex & 916 \\
			Citibank & 1852		 & Qualcomm & 919 \\
			Colgate-Palmolive & 776 	 & Quest Global & 	150 \\
			Creative Technology & 150     & Rackspace Hosting & 732 \\
			Darden Restaurants & 994 	 & Samsung & 150 \\
			DBS & 288 			 & SCB & 942 \\
			Devon Energy & 336 		 & Singtel & 480 \\
			DreamWorks Animation & 978 & StarBucks & 828 \\
			EOG Resources & 127 	 & Starhub & 150 \\
			FactSet & 976 & Stryker 	 & 904 \\
			FedEx Corporation & 850 	 & SVB Financial Software & 277 \\
			Flextronics & 150 		 & J.M. Smucker Company & 318 \\
			General Mills & 1036 		 & Ultimate Software & 524 \\
			Goldman Sachs & 970 	 & Umpqua Bank & 242 \\
			Google & 756 			 & Union Overseas Bank & 265 \\
			Hasbro & 458 			 & World Foods Market & 757 \\
			Herman Miller & 540 		 & Yes Bank & 176 \\
			\hline
			\end{tabular}
            \end{scriptsize}
			\vspace{5mm}
			\caption{Number of reviews per company}
			\label{table:freqc}
	\end{table}
            
	In Table \ref{table:freqc}, we show the number of reviews per company. The aim was to collect 500-1000 most helpful reviews\footnote{Reviews for each company in the dataset were selected based on them bearing the most `helpful' review tag provided by \url{Glassdoor.com}}. However, for some companies the number of reviews available on Glassdoor numbered less than 500.


	\subsection{Preprocessing}
	
	As mentioned earlier, the reviews follow a rigid outline structure regardless of writers. Thus, we shuffled the dataset using a pseudo-random generator so as to break any kind of patterns embedded in the dataset. For the purposes of processing the text, we removed any urls, links and hashtags with the use of regular expressions. However, we retained the smileys and emoticons used in the reviews and included them in our vocabulary so as to exploit the emotional and sentimental hints present in them. As we used a context-based algorithm to create our aspect-sentiment embeddings, retaining these special, non-verbal `words' held a key importance in the performance of the ELM classification. Following this, we used the \textit{NLTK Tokenize Package} to tokenize the reviews into sentences and, finally, into words so as to build up our model's vocabulary.
	
	\section{Backgrounds}
	\label{sec:basic}
	\subsection{Aspect-sentiment Embeddings}
	In this paper, we introduce Aspect-sentiment Embeddings, which projects companies onto an n-dimensional space. In particular, each company is given a sentiment strength for each aspect (e.g., salary, work life, location) based on the opinions mined from employee reviews of the company. In mathematical notation we can say, $(s_1, s_2,.....,s_n)$ is a vector where $s_i$ is the sentiment score for aspect $i$ and there are a total of $n$ aspects. We constructed such vectors for every company in our dataset, which gave us aspect-sentiment embeddings of the companies.  
	\subsection{Doc2vec for review level embeddings} 
	
	As we use ensemble architecture to prepare our aspect-sentiment embeddings, the ELM module plays a crucial role as one of the dual paths in the architectural model. To use the ELM module, we need to convert the raw text into review-level summarized embeddings. In our work, we use \textit{Doc2vec}~\cite{le2014distributed} to achieve this task. Doc2vec, also known as paragraph2vec, is a modification of the word2vec algorithm. Word2vec itself is a famous algorithm for word embeddings provided by \cite{mikolov2013efficient} which trains a neural network to extract contextual-based word embeddings based on the CBOW architecture. Such training has been done on a 100 billion words corpus from Google News and the vectors formed are of 300 dimensionality. In contrast, Doc2vec is an unsupervised learning of continuous representations for larger blocks of text, such as sentences, paragraphs or entire documents. As the reviews in our dataset are very detailed, employing the Doc2vec algorithm is justified. We used the python implementation provided by \textit{gensim}\footnote{https://radimrehurek.com/gensim/} to extract 300 dimensional embeddings to be fed into the ELM model for sentimental analysis and classification.

	\subsection{Sentiment Dictionary/Lexicons}
	
	In order to assign aspect polarity score, we created a dictionary of terms along with their polarities to be used by our ensemble algorithm. We used two primary resources, SentiWordNet \cite{esuli2006SentiWordNet} and SenticNet \cite{camnt4} to create this dictionary. To filter out irrelevant words that act as noise and would thus fail to provide good polarity to the aspects, we kept an absolute threshold of 0.25 in the polarity scores of the terms in both resources. In order to avoid redundancy, when a particular term is present in both SentiWordNet and SenticNet, we gave priority to SentiWordNet and chose the term polarity pair from there. Once the dictionary was created, we used it as a reference lookup table in our algorithm mentioned below.
	\subsection{Aspect Level Sentiment Analysis}
	In opinion mining, different levels of analysis granularity have been proposed, with each having its own advantages and drawbacks \cite{cambria2013stat}. Aspect-based opinion mining \cite{hu2004mining} and \cite{ding2008holistic} focuses on the relations between aspects and document polarity. An aspect, also known as an opinion target, is a concept in which the opinion is expressed in the given document. For example, the sentence, ``The screen of my phone is really nice and its resolution is superb'' contains positive polarity for a phone review, i.e., the author likes the phone. However, more specifically, the positive opinion is about its screen and resolution; these concepts are called opinion targets, or, aspects of this opinion. The task of identifying the aspects in a given opinionated text is called aspect extraction.
	
	There are two types of aspects defined in aspect-based opinion mining: explicit aspects and implicit aspects. Explicit aspects are words in the opinionated document that explicitly denote the opinion target. For instance, in the aforementioned example, the opinion targets `screen' and `resolution' are explicitly mentioned in the text. In contrast, an implicit aspect is a concept that represents the opinion target of an opinionated document, but which is not specified explicitly in the text. One can infer that the sentence, ``This camera is sleek and very affordable'' implicitly contains a positive opinion of the aspects `appearance' and `price' of the entity camera. These same aspects would be explicit in an equivalent sentence: ``The appearance of this camera is sleek and its price is very affordable''.
	
    \begin{figure}[t] 
      \centering 
      \includegraphics[width = .99\textwidth]{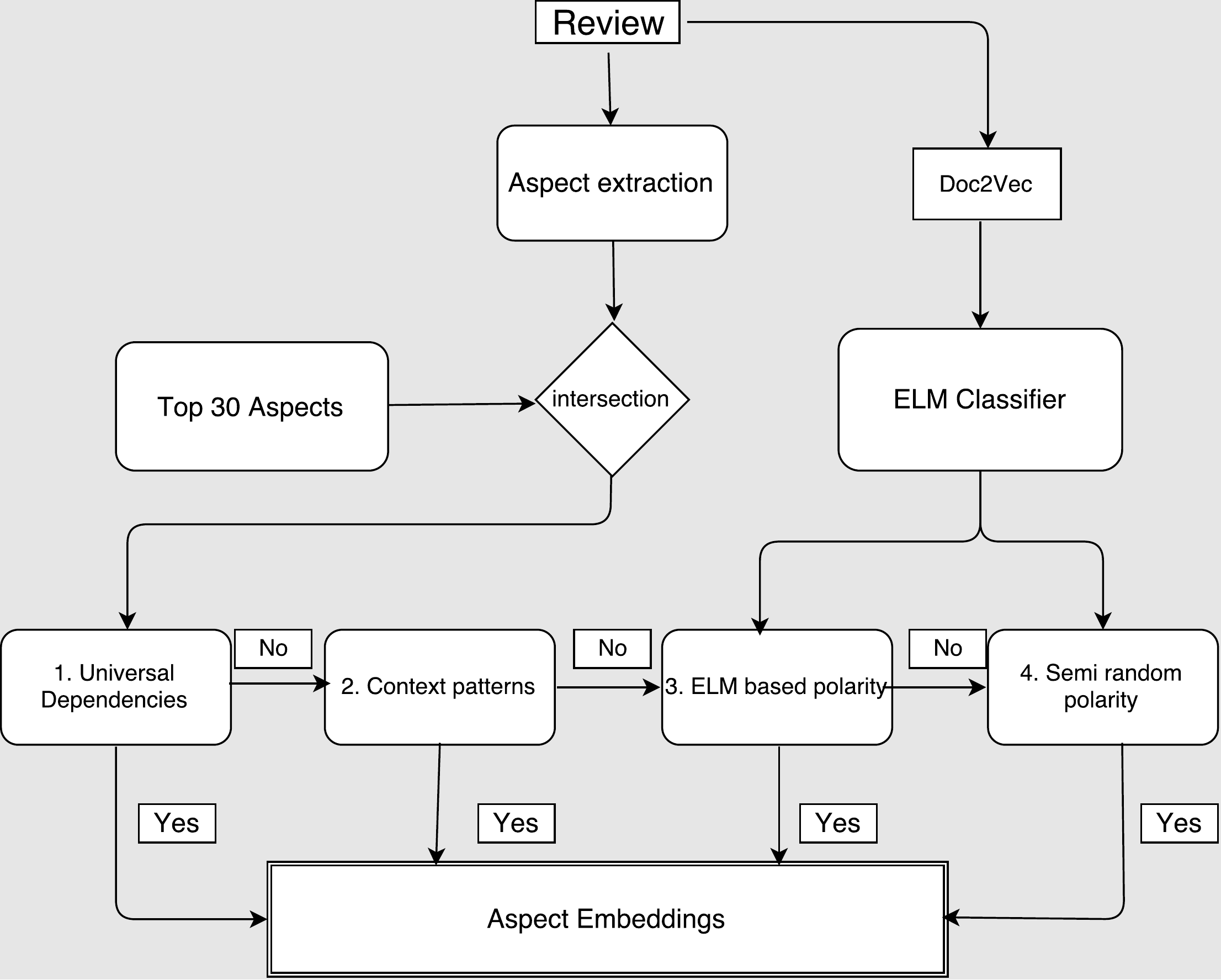}   
      \caption[]{The flowchart of the algorithm.}
      \label{fig:flowchart}
	\end{figure}
    
	\subsection{Extreme learning machine (ELM)}
	\label{elmo}
	For classification, we used ELM as a supervised classifier. The ELM approach \cite{Huang2011} was introduced to overcome some issues in back-propagation network \cite{Ridella1997} training, specifically, potentially slow convergence rates, the critical tuning of optimization parameters \cite{Vogl1988}, and the presence of local minima that call for multi-start and re-training strategies. The ELM learning problem settings require a training set, $X$, of $N$ labeled pairs, using the equation $(\mathbf{x}_i, y_i)$, where $\mathbf{x}_i \in \mathcal{R}^m$ is the $i$-th input vector and $y_i \in \mathcal{R}$ is the associate expected `target' value; using a scalar output implies that the network has one output unit, without loss of generality.
	
	The input layer has $m$ neurons and connects to the `hidden' layer (having $N_h$ neurons) through a set of weights $\{\hat{\mathbf{w}}_j \in \mathcal{R}^m; j=1,...,N_h\}$. The $j$-th hidden neuron embeds a bias term, $\hat{b}_j$,and a nonlinear `activation' function, $\varphi(\cdot)$; thus the neuron's response to an input stimulus, $\mathbf{x}$, is:
	
	\begin{equation}
	\label{act_f}
	a_j(\mathbf{x}) = \varphi(\hat{\mathbf{w}}_j \cdot \mathbf{x} + \hat{b}_j)
	\end{equation}
	
	Note that (\ref{act_f}) can be further generalized to a wider class of functions \cite{Huang2006} but for the subsequent analysis this aspect is not relevant. A vector of weighted links, $\bar{\mathbf{w}}_j \in \mathcal{R}^{N_h}$, connects hidden neurons to the output neuron without any bias \cite{huains2}. The overall output function, $f(\mathbf{x})$, of the network is:
	
	\begin{equation}
	\label{output_f}
	f(\mathbf{x}) = \sum_{j=1}^{N_h} \bar{\mathbf{w}}_j a_j(\mathbf{x})
	\end{equation}
	
	It is convenient to define an `activation matrix', $\mathbf{H}$, such that the entry $\{h_{ij} \in \mathbf{H}; i=1,...,N; j=1,...,N_h\}$ is the activation value of the $j$-th hidden neuron for the $i$-th input pattern. The $\mathbf{H}$ matrix is:
	
	\begin{equation}
	\label{h_matrix}
	\mathbf{H} \equiv 
	\begin{bmatrix} \varphi(\hat{\mathbf{w}}_1 \cdot \mathbf{x}_1 + \hat{b}_1) & \cdots & \varphi(\hat{\mathbf{w}}_{N_h} \cdot \mathbf{x}_1 + \hat{b}_{N_h}) \\ \vdots &
	\ddots & \vdots \\ \varphi(\hat{\mathbf{w}}_1 \cdot \mathbf{x}_N + \hat{b}_1) & \cdots & \varphi(\hat{\mathbf{w}}_{N_h} \cdot \mathbf{x}_N + \hat{b}_{N_h}) \end{bmatrix}
	\end{equation}
	
	In the ELM model, the quantities $\{\hat{\mathbf{w}}_j, \hat{b}_j\}$ in (\ref{act_f}) are set randomly and are not subject to any adjustment, and the quantities $\{\bar{\mathbf{w}}_j, \bar{b}\}$ in (\ref{output_f}) are the only degrees of freedom. The training problem reduces to the minimization of the convex cost:
	
	\begin{equation}
	\label{min}
	\min_{\{ \bar{\mathbf{w}}, \bar{b} \}} {\begin{Vmatrix} \mathbf{H} \bar{\mathbf{w}} - \mathbf{y} \end{Vmatrix}}^2
	\end{equation}
	
	A matrix pseudo-inversion yields the unique $L_2$ solution, as proven in \cite{Huang2011}:
	
	\begin{equation}
	\label{solution}
	\bar{\mathbf{w}} = \mathbf{H}^+ \mathbf{y}
	\end{equation}
	
	The simple, efficient procedure to train an ELM therefore involves the following steps:
	
	\begin{enumerate}
		\item Randomly set the input weights $\hat{\mathbf{w}}_i$ and bias $\hat{b}_i$ for each hidden neuron;
		
		\item Compute the activation matrix, $\mathbf{H}$, as per (\ref{h_matrix});
		
		\item Compute the output weights by solving a pseudo-inverse problem as per (\ref{solution}).
	\end{enumerate}
	
	Despite the apparent simplicity of the ELM approach, the crucial result is that even random weights in the hidden layer endow a network with a notable representation ability \cite{Huang2011}. Moreover, the theory derived in \cite{Huang2012} proves that regularization strategies can further improve its generalization performance. As a result, the cost function (\ref{min}) is augmented by an $L_2$ regularization factor as follows:
	
	\begin{equation}
	\label{reg}
	\min_{\bar{\mathbf{w}}} {\{{\begin{Vmatrix} \mathbf{H} \bar{\mathbf{w}} - \mathbf{y} \end{Vmatrix}}^2 + \lambda {\begin{Vmatrix} \bar{\mathbf{w}} \end{Vmatrix}}^2\}}
	\end{equation}
	
	\section{Detailed Algorithm: Ensemble Architecture}
	\label{sec:algo}
	We propose a hybrid algorithm, which works as an ensemble of Unsupervised and Machine Learning approaches, for assigning sentiment labels to the reviews. First, based on the dependency structure of a review we assign polarity to the aspects present in the reviews. This process assumes that the aspect word is connected to a word that is polar and present in the sentiment lexicon. If there is no polar word found connected to the aspect word, according to the sentiment dictionary, we resort to the use of supervised classifier, i.e., ELM. The usage of ELM has multiple benefits like comparable or better performance than other machine learning models like SVMs, and most importantly boasts a significant reduction in model building time. Training time is an important aspect in our work given its high probability to be adapted into an online and real-time application. The flowchart of the proposed algorithm is shown in Figure \ref{fig:flowchart}.
	
	\subsection{Aspect Extraction}
	We used the aspect extraction method by Poria et al.~\cite{poria2016aspect}, who proposed a hybrid classifier that uses a Convolutional Neural Network for aspect extraction, as well as linguistic patterns for the purposes of pruning the aspect extraction process. The extracted aspects are given below:
	
	\begin{itemize}
		
		\item \texttt{Job}, is an umbrella aspect that represents the overall characteristic and nature of the job at that particular company, e.g., \textit{Stable and secure job, good people}  
		
		\item \texttt{Employees/Co-workers}, represents the quality of employees, co workers and the company's relationship with them, e.g., \textit{Very healthy organization with a high-performance culture and very talented employees.}
		
		\item \texttt{Working time}, clubs aspects of diverse meanings ranging from `extra-time' to `time-off' which signify work hours and trends, e.g., \textit{Great people, very generous vacation and time off including sabbaticals every 5 years.}  
		
		\item \texttt{Management}, explores the managerial aspects of the company, e.g., \textit{Great place to work. Inclusive management process. Great products.}  
		
		\item \texttt{Office culture}, summarizes the office environment and the working style, e.g., \textit{Strong focus on procedures, policies, culture, and people. Great benefits.}  
		
		\item \texttt{Location}, represents the comments on location of the company, e.g., \textit{Competitive salary, Nice location, Full freedom.} 
		
		\item \texttt{Work life}, speaks in particular about the type and quality of work, along with the work-life balance present in the company, e.g., \textit{Great office space and location, interesting products to work on.} 
		
		\item \texttt{Salary}, provides information on the salary margins of the company, e.g., \textit{Salary is OK Bonus is good unless there is a food fight. Too laid back (leads to no innovation).} 
		
		\item \texttt{Perks/Benefits}, compensations, bonuses and miscellaneous benefits are included in this aspect, e.g., \textit{Good perks for this type of job - and vary even across levels of employment. Fitness reimbursement, stock options, sabbaticals, etc.} 
		
		\item \texttt{Job opportunities}, scope of the job in the company, e.g., \textit{Strong culture, good reputation, interesting opportunities, management cares about the careers of the employees they are managing.} 
		
		\item \texttt{Employee experience}, lists the sentiments of employees with different degrees of experience and also the quality of experience to be acquired in the company, e.g., \textit{The size of the org can make it difficult for individuals to have their voices heard, especially for new hires, regardless of their experience.} 
		
		\item \texttt{Official staff}, talks about the strength, quality and hospitality of the staff in the company, e.g., \textit{Hours, upper management, lack of staff.} 
		
		\item \texttt{Job training}, expresses the training frameworks and opportunities provided by the company, e.g., \textit{Inconsistent hours, sometimes no hours. No proper training.}
		
		\item \texttt{Personal growth}, the possibility of technological and experiential growth for the employee, e.g., \textit{Need patience to sense growth since it is a challenging business and changing company.}
		
		\item \texttt{Leadership}, discusses the role and efficacy of the leadership/senior officials in the areas of motivation and leadership skills. \textit{Leadership does not know how to utilize experienced, professional talent}

		\item \texttt{Politics}, represents the interaction between people for power, e.g., \textit{Politics drive people for more power.}  
		
		\item \texttt{Company business}, explores the business aspects such as performance, trade, etc. of the company, e.g., \textit{Business of the company is booming.}
		
		\item \texttt{Career development}, forecasts the future of the individuals and company, e.g., \textit{International job within 2 years.}  
		
		\item \texttt{Vacation}, represents the number of holidays the company provides its staff with, e.g., \textit{Paid vacation for the summer. Free travel within the same country.}  
		
		\item \texttt{Company support}, summarizes the quality of interaction between employees, e.g., \textit{Supervisors take care of their employees. }  
		
		\item \texttt{Flexibility}, represents how flexible the company's environment is, e.g., \textit{Full freedom, work from home allowed.} 
		
		\item \texttt{Performance}, speaks about the type and quality of work done by the employees, e.g., \textit{Extraordinary skills shown by the employees.} 
		
		\item \texttt{Job respect}, shows how employees admire their peers and supervisors, e.g., \textit{Mutual respect amongst the employees.} 
		
		\item \texttt{Work projects}, companies projects, products and plans are included in this aspect, e.g., \textit{Diverse products,numerous projects are provided by the company.} 
		
		\item \texttt{Market viability}, provides information about the type and quality of the market, the company battles, e.g., \textit{Competitive, changing market.} 
		
		\item \texttt{Technology}, explores the technological aspects of the company, e.g., \textit{The machinery used in this company is built on ancient technology.} 
		
		\item \texttt{Work issues}, highlights the operational, legal and internal issues of the company, e.g., \textit{Most of the machines are not working.} 
		
		\item \texttt{Knowledge scope}, lists skills required by the company and knowledge acquired by the employees, e.g., \textit{Technical knowledge in required for this task.}
		
		\item \texttt{Employee communication}, discusses the interaction between employees and the companies, e.g., \textit{People are very interactive in this company.}
		
		\item \texttt{Stress}, stress and pressure the employees and companies feel, e.g.,  \textit{Getting underpaid, stress good for working in a competitive environment.}
		
		\end{itemize}
We also show the corpus frequency of these aspects in Table \ref{table:freqa}.

    \begin{table}
      \centering
      \begin{tabular}{|c|c|c|c|} 
          \hline
          Aspect &  Frequency & Aspect &  Frequency   \\ [0.5ex] 
          \hline\hline
          Employees/Co-workers & 7659 & Employee experience & 1288 \\
          Work Life & 7305 & Location & 627	\\
          Perks/Benefits & 4565 & Leadership & 599	\\
          Office culture & 4192 &  Technology & 490	\\
          Working time & 3658  & 	Politics & 451	\\
          Salary & 3323 & Flexibility & 340 \\
          Management & 2654 & Company business & 116	\\
          Job opportunities & 2129 &&						\\
          \hline
      \end{tabular}
      \vspace{5mm}
      \caption{Extracted aspects with their corpus frequency.}
      \vspace{-5mm}
      \label{table:freqa}
	\end{table}
    
	\subsection{Assigning Polarity to the Aspects}
	\label{sec:pol}
	
	In this section, we describe the process of assigning polarity to the aspects.	
	\subsubsection{Universal Dependent Modifiers}
	
	Keeping in mind the goal of finding the polarity score of each aspect (out of the top 30 extracted aspects) present in a review, we start with finding the universal dependencies\footnote{http://universaldependencies.org/} of aspects in the review. We focus primarily on three dependencies, namely, \textit{adjectival modifier (\texttt{amod}), adverbial modifier (\texttt{advmod}) and nominal subject (\texttt{nsubj}).}
	
	For better understanding, we provide some examples from reviews in our dataset, with which we demonstrate the aforementioned dependencies associated with the aspects present.

	\begin{itemize}
		\item \textit{Great opportunities for career growth.}
		\texttt{amod}(opportunities, Great)
		\item \textit{Very political and conservative company. Old school, stodgy.}
		\texttt{advmod}(political, Very)
		\item \textit{Great people to work with, perks of business traveling.}
		\texttt{nsubj}(travelling, perks) 
	\end{itemize}
	
	We use \textit{StanfordCoreNLP Parser} as the tool to extract these universal dependencies. After we find the dependencies, we use these `trigger' words to determine the sentimental polarity of the corresponding aspect. As the aspect sentiment is determined by these modifiers, we lookup their polarities in the prepared sentiment dictionary. These polarities serve as the corresponding aspect score for that particular aspect. This process is repeated for the top 30 aspects that are found in the review.
	
	\subsubsection{Context patterns}
	
	In the event that the trigger word is not in the sentiment dictionary, we move on to the second step in the ensemble. Here, we look at the context (window size - 5 words used, including the aspect). We try to find dependency patterns mentioned by \cite{poria2014dependency} and use them to determine the overall polarity score for the aspect. Our assumption is that the presence of highly polar words in the context will contribute to the overall polarity of the aspect. Negations have been appropriately handled as they flip the polarity. 
	
	\subsubsection{ELM based polarity score}
	
	Should the two procedures mentioned above fail to assign a score to the aspect, we use the prediction made by our ELM model ( 1 - positive,  0 - negative). We directly lookup the polarity of the aspect word in the sentiment dictionary and adjust it based on the ELM output as per the following formula:
	\[
	aspect_{score} = (e_{out})*(lookup(aspect)) + (1-e_{out})*( -1 * lookup(aspect))
	\] 
	
	here, \textit{e\_out} is the output predicted by the ELM for the review, \textit{lookup(aspect)} is the polarity score of the aspect word as obtained from the sentiment dictionary. 
	
	\subsubsection{ELM based semi random score}
	
	If the aspect word itself is not present in the sentiment dictionary, we initialize a random polarity value based on the ELM output. The following formula is used to generate the random score: 
	
	\[
	aspect_{score} = 
	\begin{cases} 
	rand(0,1) & , e_{out} \equiv 1 \\
	rand(-1,0) & , e_{out} \equiv 0 \\
	
	\end{cases}
	\]
	
	Here, \textit{rand(a,b)} is a random generator function which generates random real numbers within the range [a,b].
	
	We had to assign a score randomly to the aspects in only 2\% of cases.  This indicates that the semi-random polarity generation of the aspects did not impact the overall aspect-sentiment embeddings much. However, a fully automatic process is always desirable and as such, we plan on doing so in our future work.
    
	\section{Experimental Results}
	\label{sec:exp}
In this section, we describe the experimental results and insights drawn from the crawled datasets. 
Table \ref{table:aspects} (Appendix~\ref{sec:topaspects}) shows the aspects and aspect terms that we extracted from the dataset.

We utilized both SVM and ELM models (Table \ref{table:svmelm}) on the dataset in order to detect sentiment. In the experiments, the SVM method performed better than ELM in terms of accuracy. However, the difference between the performances of these classifiers on the given dataset was not statistically significant, as per the paired t-test (p>0.05). Also, in the case of training time, we observed that the ELM method was almost 30 times faster than that of the SVM method on this dataset.

	\begin{table}[h]
		\centering
		\begin{tabular}{|c| c | c|} 
			\hline
			Model &  Accuracy & Macro F1-score  \\ [0.5ex] 
			\hline\hline
			SVM & 75.13\%   &   74.85\% \\
			ELM & 74.89\%   &   75.09\% \\
			\hline
		\end{tabular}
		\vspace{5mm}
		\caption{Performance of SVM and ELM on the dataset.}
		\label{table:svmelm}
	\end{table}

    \begin{figure}[h] 
      \centering 
      \includegraphics[width = .99\textwidth]{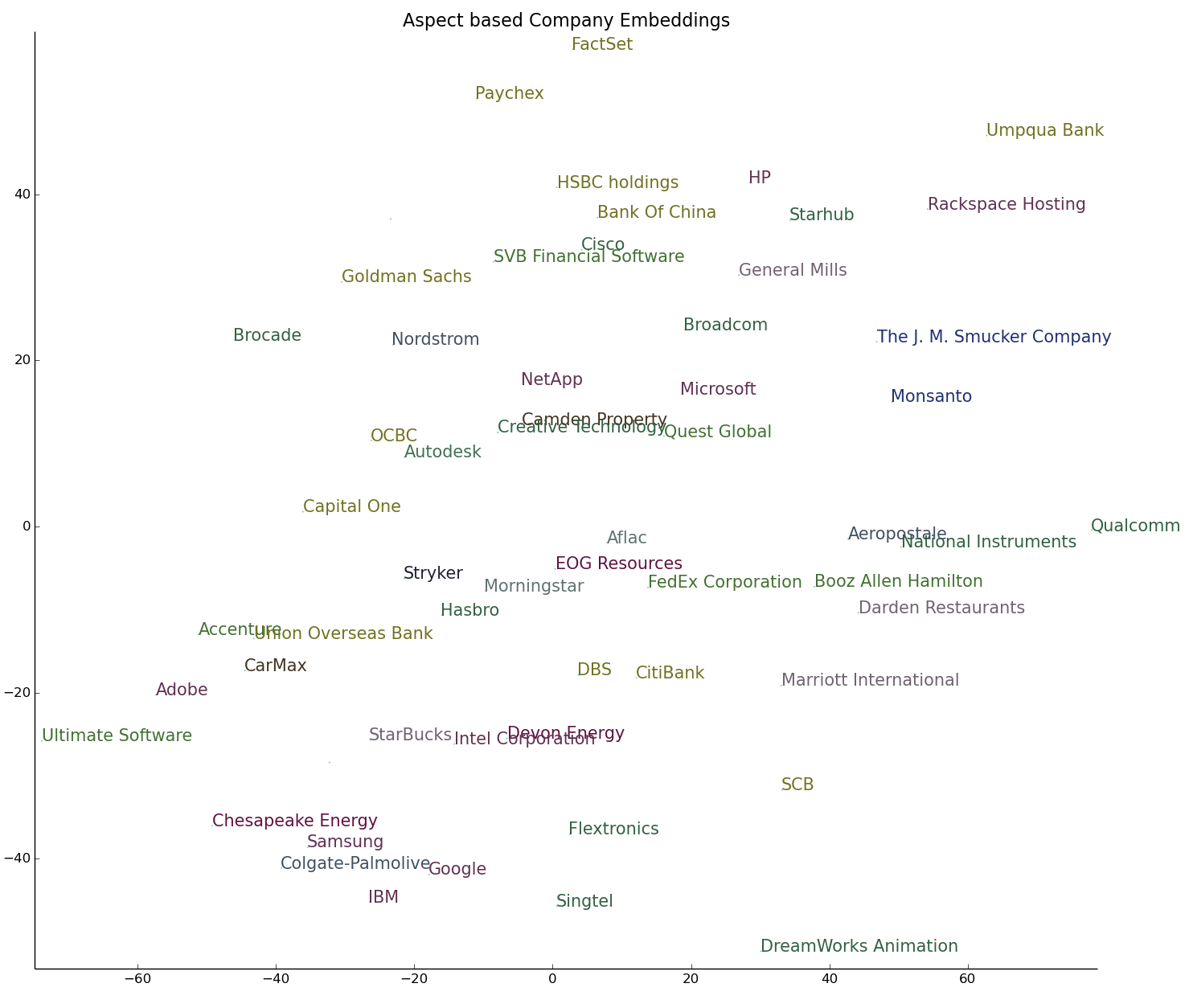}   
      \caption[]{Projection of the Aspect-sentiment Embeddings of the companies. Note: The same color represents companies from the same sector.}
      \label{ce.png}
    \end{figure}
	
	\subsection{Aspect-sentiment Embeddings}
In Section \ref{sec:pol}, we described the process of assigning polarity to the aspects. We then calculated the score of an aspect belonging to a company by simply taking the average score of the polarities belonging to that aspect in all reviews of said company.

\begin{table}[!htbp]
\centering
	\begin{tabular}{|c| c| c| c |} 
		\hline
		Company 1 &  Company 2 &   Cosine Similarity  \\ [0.5ex] 
		\hline\hline
		Accenture & Booz Allen Hamilton &  0.548 \\
		Accenture & FedEx & 0.733\\
		\hline
		Google & Microsoft & 0.549\\
		Microsoft & Intel & 0.486\\
		Adobe & HP & 0.370\\
		Adobe & Google & 0.276\\
		Adobe & IBM & -0.214\\
		
		\hline
		OCBC & Goldman Sachs & 0.422\\
		Goldman Sachs & DBS & -0.098\\
		DBS & SCB & 0.313\\
		Goldman Sachs & SCB & 0.041\\
		\hline
		Singtel& Broadcom & 0.424\\
		Singtel & Starhub & -0.244\\
		
		\hline
		Microsoft & Stryker & 0.687\\
		National Instruments & Microsoft & -0.117\\
		NetApp & Hasbro & 0.655\\
		Monsanto & Quest GLobal & -0.380\\
		\hline
	\end{tabular}
	\vspace{5mm}
		\caption{Cosine Similarities between the Companies (calculated based on the aspect-sentiment embeddings)}
		\label{table:cosim}
\end{table}
 If we consider each aspect as a separate dimension, and the polarity value of a company for one aspect is the projection along that dimension, then we can project each company in a n-dimensional space where n = number of aspects. In our case, n = 30. In Figure \ref{ce.png} we show projection of the companies using aspect-sentiment embeddings.

 The motivation for constructing aspect-sentiment embeddings for the companies was to be able to calculate the similarities between companies based on the sentiments of the employees working at those companies.
   
  In Table \ref{table:cosim}, we present the cosine similarity scores between companies from similar or differing sectors. We see that even though \emph{Goldman Sachs} and \emph{DBS} are in same sector, i.e., Banking and Finance, they have a lower similarity score. However \emph{DBS} and \emph{SCB}(Standard Chartered Bank) have a relatively higher similarity score.

 \begin{table}[!htbp]
 	\centering
 	\begin{tabular}{|c|c|c|c|c|c|c|}
 		\hline
 		&
 		\multicolumn{2}{c |}{Location} &
 		\multicolumn{2}{c |}{Salary} &
 		\multicolumn{2}{c |}{Work Life} \\
 		\cline{2-7}
 		& Tech & Finance & Tech & Finance & Tech & Finance \\
 		\hline
 		B&  \scriptsize{Microsoft}& \scriptsize{Umpqua Bank}& \scriptsize{Intel}& \scriptsize{Yes Bank}& \scriptsize{Adobe}& \scriptsize{Goldman S}\\
 		
 		E&  \scriptsize{Intel}& \scriptsize{Goldman S}& \scriptsize{Adobe}& \scriptsize{Goldman S} & \scriptsize{Microsoft}& \scriptsize{Bank of China}\\
 		S&  \scriptsize{Adobe}& \scriptsize{SCB}& \scriptsize{Microsoft}& \scriptsize{HSBC}& \scriptsize{Google}& \scriptsize{SCB}\\
 		T&  \scriptsize{Google}& \scriptsize{Citibank}& \scriptsize{Cisco}& \scriptsize{OCBC}& \scriptsize{Cisco}& \scriptsize{UOB}\\
 		&  \scriptsize{HP}& \scriptsize{HSBC}& \scriptsize{Google}& \scriptsize{Citibank}& \scriptsize{FactSet}& \scriptsize{HSBC}\\
 		\cline{1-7}
 		W&  \scriptsize{IBM}& \scriptsize{Yes Bank}& \scriptsize{Creative}& \scriptsize{UOB}& \scriptsize{Samsung}& \scriptsize{Citibank}\\
 		O&  \scriptsize{Creative}& \scriptsize{OCBC}& \scriptsize{Flextronics}& \scriptsize{Bank of China}& \scriptsize{HP}& \scriptsize{DBS}\\
 		R&  \scriptsize{NetApp}& \scriptsize{DBS}& \scriptsize{FactSet}& \scriptsize{SCB}& \scriptsize{NetApp}& \scriptsize{OCBC}\\
 		S&  \scriptsize{Cisco}& \scriptsize{UOB}& \scriptsize{Samsung}& \scriptsize{Umpqua Bank}& \scriptsize{Creative}& \scriptsize{Umpqua Bank}\\\
 		T&  \scriptsize{FactSet}& \scriptsize{Bank of China}& \scriptsize{HP}& \scriptsize{DBS}& \scriptsize{Intuit}& \scriptsize{Yes Bank}\\
 		\hline
 	\end{tabular}
 	\vspace{5mm}
 	\caption{Companies with best/worst salary and work culture rating in tech and finance sectors.}
 	\label{table:comp}
 \end{table}
 
\begin{figure*}[t]
    \centering
    \begin{subfigure}[t]{0.5\textwidth}
        \centering
        \includegraphics[scale=0.4]{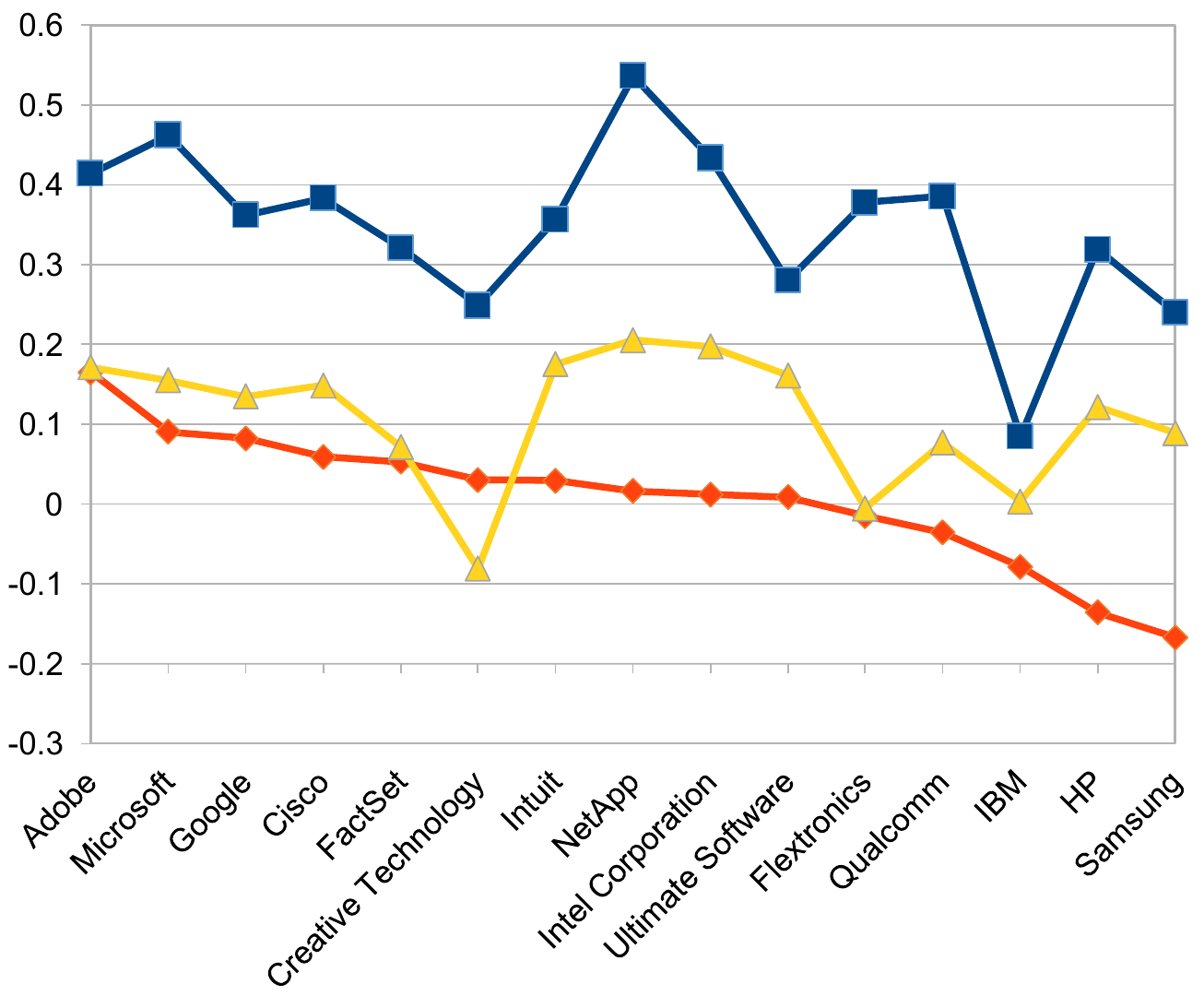}
        \caption{Tech sector.}
        \label{tech.pdf}
    \end{subfigure}%
    ~ 
    \begin{subfigure}[t]{0.5\textwidth}
        \centering
        \includegraphics[scale=0.36]{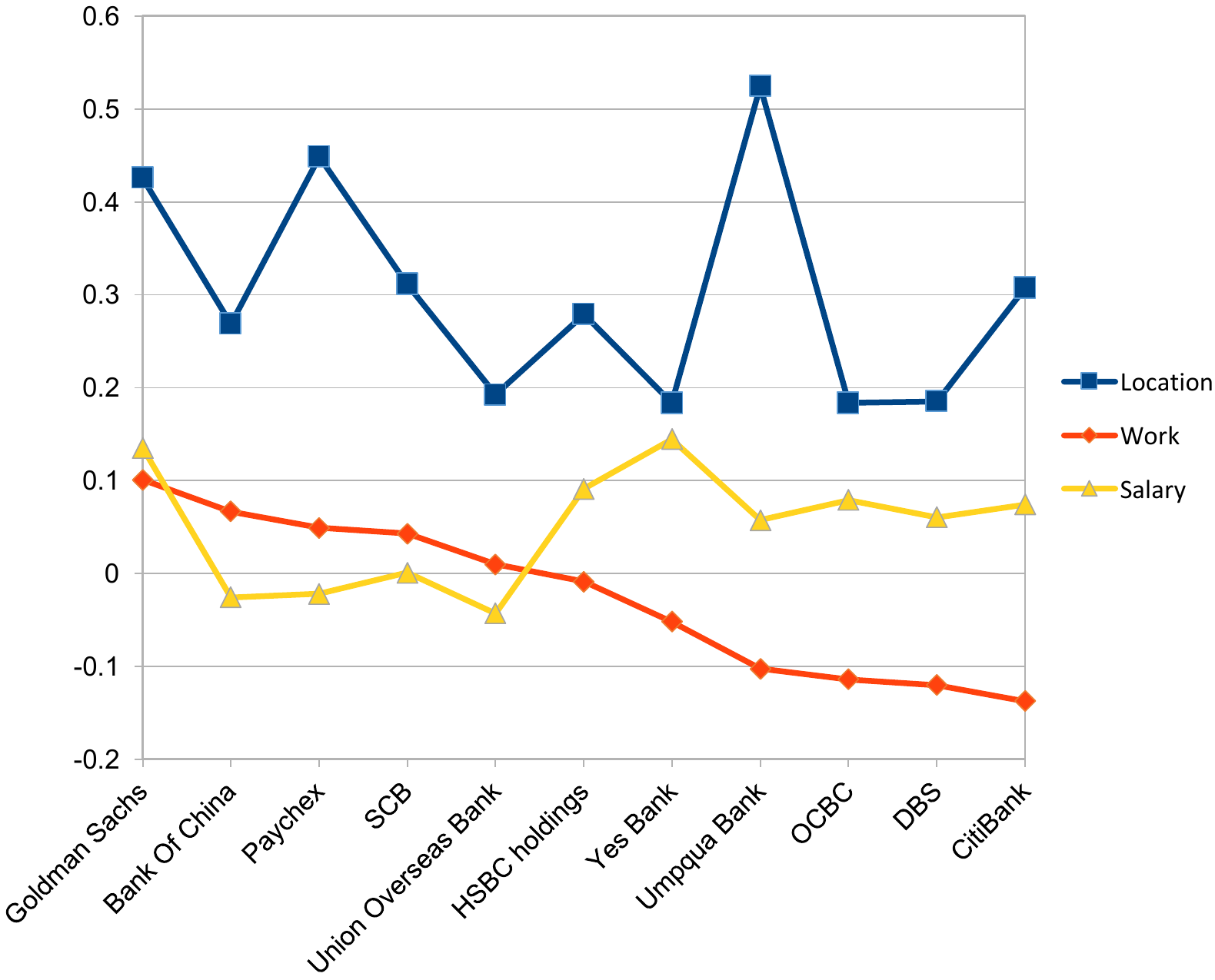}
        \caption{Finance sector.}
        \label{finance.pdf}
    \end{subfigure}
    \caption{The plot of the polarity of the aspects of companies in tech and finance sector.}
\end{figure*}

  Table \ref{table:comp} presents the best and worst companies in the technological and finance sectors based on sentiment values of the salary, location and work life aspects. Analysis (Figure \ref{finance.pdf} and \ref{tech.pdf}) shows that employees are mostly happy with the salary they receive in both the finance and tech sectors. However, in relation to most banks, employees provide negative feedback on work culture. And, although tech companies receive positive feedback for their work culture, the intensity of such positivity is comparatively lower than salary satisfaction.
 
 \section{Conclusion}
 \label{sec:con}
In this paper, we described the process of constructing aspect-sentiment embeddings of companies. In particular, we employed aspect-level sentiment analysis on the previously barely researched employee reviews of various companies available on the site Glassdoor. Several experimental insights of the data are given in this study. We addressed the overall employee sentiment on different aspect granularities, e.g., \emph{salary}, \emph{location}, \emph{work life}, etc. This study presents a useful tool for companies to address employees' concerns and increase staff morale. On the other hand, job seekers will also be able to use this study to better find the best employers in their domain of interests. 

Future work will mainly focus on considering the rating already given by the users in order to develop a user-product-sentiment model. A more comprehensive aspect-level sentiment analysis is also an important part of this future work.

\section*{Acknowledgment}
This research was supported in part by the National Natural Science Foundation
of China under Grant no.~61472266 and by the National University of Singapore
(Suzhou) Research Institute, 377 Lin Quan Street, Suzhou Industrial Park, Jiang
Su, People's Republic of China, 215123.

\newpage
 \bibliography{refs-AG}
 \bibliographystyle{plain}

\appendix
\newpage
\section{Top aspects and their aspect-terms} \label{sec:topaspects}

	\begin{table}[!htbp]
		\centering
		\begin{tabular}{|c|c|c|c|}
			\hline
			\multirow{2}{*}{\textbf{Aspects}} & \multirow{2}{*}{\textbf{Aspect Terms}} & \multirow{2}{*}{\textbf{Aspects}} & \multirow{2}{*}{\textbf{Aspect Terms}}\\
			&&&\\
			\hline
            \footnotesize{Company}& \scriptsize{Professional, Booming, Structured}& \footnotesize{Career}& \scriptsize{Rewarding, International, Guided}\\
            \footnotesize{business}&\scriptsize{Challenging, Competitive, Steady}&\footnotesize{development}&\scriptsize{Challenging, Difficult, Solid}\\ \hline
            \footnotesize{Employee}&\scriptsize{Strong, Transparent, Cryptic,}&\footnotesize{Office}&\scriptsize{Cooperative, Balanced, Exciting,}\\
            \footnotesize{communication}&\scriptsize{Remote, Awful, Effective}&\footnotesize{culture}&\scriptsize{Suffered, Abysmal, Bias}\\ \hline
            \footnotesize{Employees}&\scriptsize{Excellent, Cooperative, Competent,}&\footnotesize{Employee}&\scriptsize{Diverse, Useful, Firsthand,}\\
            \footnotesize{/Co-workers}&\scriptsize{Stagnant, Unfriendly, Pretend}&\footnotesize{experience}&\scriptsize{Horrific, Odd, International}\\ \hline
            \footnotesize{Flexibility}&\scriptsize{Strict, Dependent, Tremendous,}&\footnotesize{Personal}&\scriptsize{Exponential, Poor, Constant,}\\
            \footnotesize{}&\scriptsize{Encourage, Minimal, Great}&\footnotesize{growth}&\scriptsize{Hierarchy, Potential, Constrain}\\ \hline
            \footnotesize{Work}&\scriptsize{Legal, Serious, Inherent,}&\footnotesize{Overall}&\scriptsize{Excellent, Temporary, Overnight,}\\
            \footnotesize{issues}&\scriptsize{Internal, Operational, Demographic}&\footnotesize{job}&\scriptsize{Changing, Tough, Secure}\\ \hline
            \footnotesize{Knowledge}&\scriptsize{Immense, Required, Sharing,}&\footnotesize{Leadership}&\scriptsize{Appreciate, Strong, Poor,}\\
            \footnotesize{scope}&\scriptsize{Technical, Limited, Vast}&\footnotesize{}&\scriptsize{Unwilling, Dedicated, Driving}\\ \hline
            \footnotesize{Location}&\scriptsize{Strategic, Remote, Accessible,}&\footnotesize{Management}&\scriptsize{Flexible, Fluctuate, Inexperienced,}\\
            \footnotesize{}&\scriptsize{Attractive, Multiple, Uncertain}&\footnotesize{}&\scriptsize{Mindful, Dishonest, Focus}\\ \hline
            \footnotesize{Market}&\scriptsize{Changing, Shrinking, Impact,}&\footnotesize{Job}&\scriptsize{Excellent, Driven, Mindset,}\\
            \footnotesize{viability}&\scriptsize{Competitive, Unknown, Successful}&\footnotesize{opportunities}&\scriptsize{International, Lacking, Unique}\\ \hline
            \footnotesize{Perks/Benefits}&\scriptsize{Unique, Scares, Incredible,}&\footnotesize{Performance}&\scriptsize{Personal, Necessary, Measurable,}\\
            \footnotesize{}&\scriptsize{Illusion, Lousy, Incentives}&\footnotesize{}&\scriptsize{Extraordinary, Encouraged, Technical}\\ \hline
            \footnotesize{Politics}&\scriptsize{Dysfunctional, Drive, Dirty,}&\footnotesize{Work}&\scriptsize{Numerous, Diverse, Challenging,}\\
            \footnotesize{}&\scriptsize{Extreme, Everywhere, Internal}&\footnotesize{projects}&\scriptsize{Pushed, Creative, Unbearable}\\ \hline
            \footnotesize{Job respect}&\scriptsize{Professional, Mutual, Utmost,}&\footnotesize{Salary}&\scriptsize{Optimal, Advancement, Hikes,}\\
            \footnotesize{}&\scriptsize{Solid, Diminishing, Great}&\footnotesize{}&\scriptsize{Midrange, Fantastic, Unattractive}\\ \hline
            
            \footnotesize{Official Staff}&\scriptsize{Understanding, Excellent, Competent,}&\footnotesize{Stress}&\scriptsize{Underpaid, Excessive, Good,}\\
            \footnotesize{}&\scriptsize{Mean, Motivated, Dysfunctional}&\footnotesize{}&\scriptsize{Constant, Additional, Incompetent}\\ \hline
            
            \footnotesize{Company}&\scriptsize{Excellent, Tedious, Benefits,}&\footnotesize{Technology}&\scriptsize{Excellent, Ancient, Global,}\\
            \footnotesize{support}&\scriptsize{Supervisors, On site, Rare}&\footnotesize{}&\scriptsize{Latest, Green, Innovative}\\ \hline
            
            \footnotesize{Working time}&\scriptsize{Exciting, Peak, Tough,}&\footnotesize{Job training}&\scriptsize{Prepares, Competent, Notch,}\\
            \footnotesize{}&\scriptsize{Stressful, Extra, Irregular}&\footnotesize{}&\scriptsize{Outstanding, Outdated, Tough}\\ \hline
            
            \footnotesize{Vacation}&\scriptsize{Decent, Mandatory, Paid,}&\footnotesize{Work Life}&\scriptsize{Competent, Mundane, Exciting,}\\
            \footnotesize{}&\scriptsize{Planned, Considering, Balance}&\footnotesize{}&\scriptsize{Friendly, Versatile, Stressful}\\ \hline
            
		\end{tabular}
		\vspace{5mm}
		\caption{Top 30 aspects along with their respective aspect terms.}
		\label{table:aspects}
	\end{table}

\end{document}